\documentclass[journal]{IEEEtran}

\usepackage{times}
\usepackage{graphicx}
\usepackage{framed,multirow}
\usepackage{amssymb}
\usepackage{latexsym}
\usepackage{amsmath}
\usepackage{mathtools}
\usepackage{cite}
\usepackage{algorithm}
\usepackage{algorithmic}
\usepackage{enumerate}
\usepackage{color}
\usepackage{setspace}
\usepackage{stfloats}
\usepackage[colorlinks]{hyperref}
\UseRawInputEncoding

\begin{document}

\title{CCR: Facial Image Editing with Continuity, Consistency and Reversibility}

\author{Nan Yang,
        Xin Luan,
        Huidi Jia,
        Zhi Han \IEEEmembership{Member, IEEE},
        and Yandong Tang$^*$ \IEEEmembership{Member, IEEE}
%\thanks{$^\star$ Authors contributed equally to this work. $^*$ Corresponding author.}
\thanks{N. Yang, X. Luan, B. J. Xia, H. D. Jia, Z. Han, and Y. D. Tang are with State Key Laboratory of Robotics, Shenyang Institute of Automation, Chinese Academy of Sciences, Shenyang 110016, China, and also Institutes for Robotics and Intelligent Manufacturing, and also University of Chinese Academy of Sciences, Beijing 100049, China(e-mail:yangnan@sia.cn; luanxin@sia.cn; jiahuidi@sia.cn; hanzhi@sia.cn; ytang@sia.cn). $^*$ Corresponding author.}}

%\markboth{IEEE TRANSACTIONS ON NEURAL NETWORKS AND LEARNING SYSTEMS, 2022}
%{Shell \MakeLowercase{\textit{et al.}}: IEEE TRANSACTIONS ON NEURAL NETWORKS AND LEARNING SYSTEMS}

% \begin{teaserfigure}
% %  \begin{center}
% %  \setcounter{figure}{0}
%     \includegraphics[width=\textwidth]{./figure_for_SPL/problem.pdf}
%     \captionof{Inconsistent and irreversible editing.}
%     \label{fig:problem}
% %   \end{center}
% \end{teaserfigure}

\maketitle

\begin{abstract}
Three problems exist in sequential facial image editing: incontinuous editing, inconsistent editing, and irreversible editing. Incontinuous editing is that the current editing can not retain the previously edited attributes. Inconsistent editing is that swapping the attribute editing orders can not yield the same results. Irreversible editing means that operating on a facial image is irreversible, especially in sequential facial image editing. In this work, we put forward three concepts and corresponding definitions: editing continuity, consistency, and reversibility. Then, we propose a novel model to achieve the goal of editing continuity, consistency, and reversibility. A sufficient criterion is defined to determine whether a model is continuous, consistent, and reversible. Extensive qualitative and quantitative experimental results validate our proposed model, and show that a continuous, consistent and reversible editing model has a more flexible editing function while preserving facial identity. Furthermore, we think that our proposed definitions and model will have wide and promising applications in multimedia processing. Code and data are available at \url{https://github.com/mickoluan/CCR}.
\end{abstract}


\IEEEpeerreviewmaketitle

\section{Introduction}
\IEEEPARstart{F}{acial} image editing has attracted extensive attention with the development of adversarial neural networks \cite{Chen2020,Zhang2020,liu2020generating,Yang2021b,shao2017collaborative}. AttGAN \cite{He2019} adopts reconstruction learning, adversarial learning, and an attribute classification constraint to achieve facial image editing. STGAN \cite{Liu2019} enhances the editing performance of AttGAN by incorporating encoder-decoder and selective transfer units. ELEGANT \cite{Xiao2018a} encodes different attributes into disentangled parts and generates images with other attributes by swapping certain parts of latent encodings. Liu et al. organize attribute labels into a hierarchical tree structure and propose a Hierarchical Style Disentanglement (HiSD) \cite{li2021} method for facial image editing. Other works \cite{Viazovetskyi2020a,Yang2021c,Yang2021a,liu2020coupled,pang2021disp+,Abdal2020a} achieve facial image editing based on a style-based architecture \cite{Karras2020,Karras2019,pang2020iterative} and some explicable semantics \cite{Shen2020,Shen2021,Upchurch2017a,Radford2015,Richardson2021}.

These facial image editing methods mentioned above and other works \cite{Zhang2019a,Zhu2019} have three common problems: incontinuous editing, inconsistent editing, and irreversible editing. An incontinuous and inconsistent editing case is shown in Fig. \ref{fig:problem}(a). We observe that: 1) the current editing can not retain the previously edited attributes, e.g., glasses; and 2) the both final editing results are inconsistent when swapping attribute editing orders. Another editing case is shown in Fig. \ref{fig:problem}(b), and we can see that the editing process is irreversible. The reason is that existing methods do not consider whether a model is continuous, consistent, and reversible.

\begin{figure*}[t]
\centering
\includegraphics[width=\textwidth]{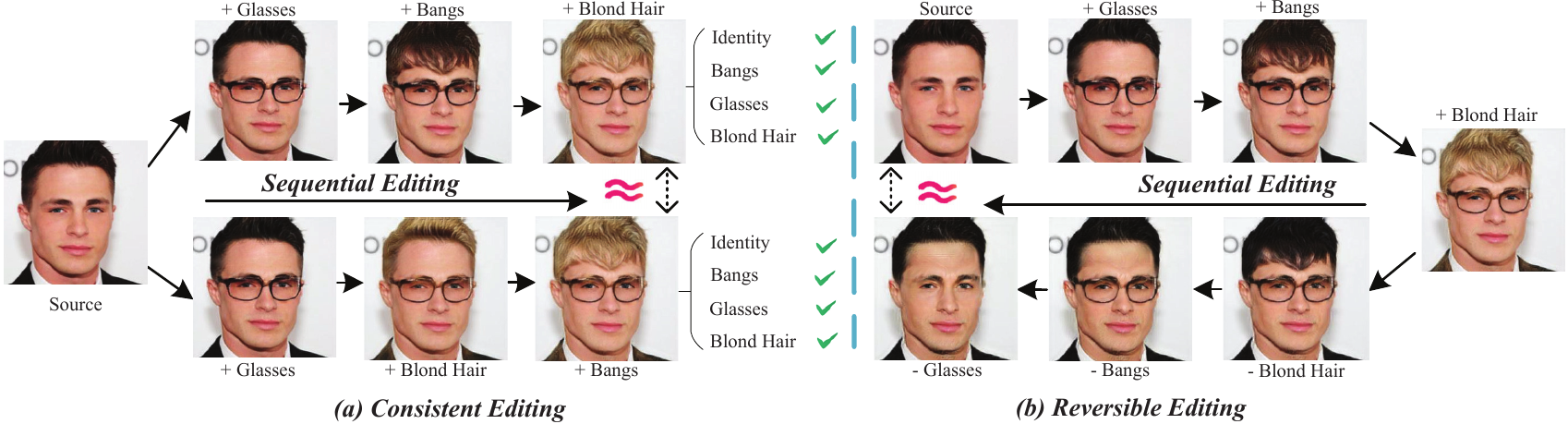}
\caption{ A case of editing continuity, consistency, and reversibility, which is obtained using our proposed model (CCR).}
\label{fig:shows}
\end{figure*}

These three problems motivate us to study a novel model, which fully considers editing continuity, consistency, and reversibility. The model should ensure that 1) the current editing can retain the previously edited attributes while preserving facial identity; 2) it can yield a consistent editing result even though swapping attribute editing orders; and 3) a facial image editing is reversible. For these purposes, we propose a continuous, consistent and reversible model to achieve the goal of editing continuity, consistency, and reversibility. The editing results of proposed model are shown in Fig. \ref{fig:shows}, and we can 1) re-edit an edited face with other attributes sequentially; 2) get a consistent editing result despite swapping attribute editing orders; and 3) obtain a reversible editing result. The results indicate that our novel model can overcome the three problems existing in current methods.

Extensive experimental results show that our proposed model achieves the goal of editing continuity, consistency, and reversibility. It also can be used to analyse disentangled attributes and the trade-off between image editing and reconstruction. We further investigate a criterion, which can be used to determine whether the model is satisfied with continuity, consistency, and reversibility. Our unique contribution that advances the field of facial image editing contains three aspects:

\begin{figure*}
    \centering
    % \captionsetup{type = figure}
    \includegraphics[width=\textwidth]{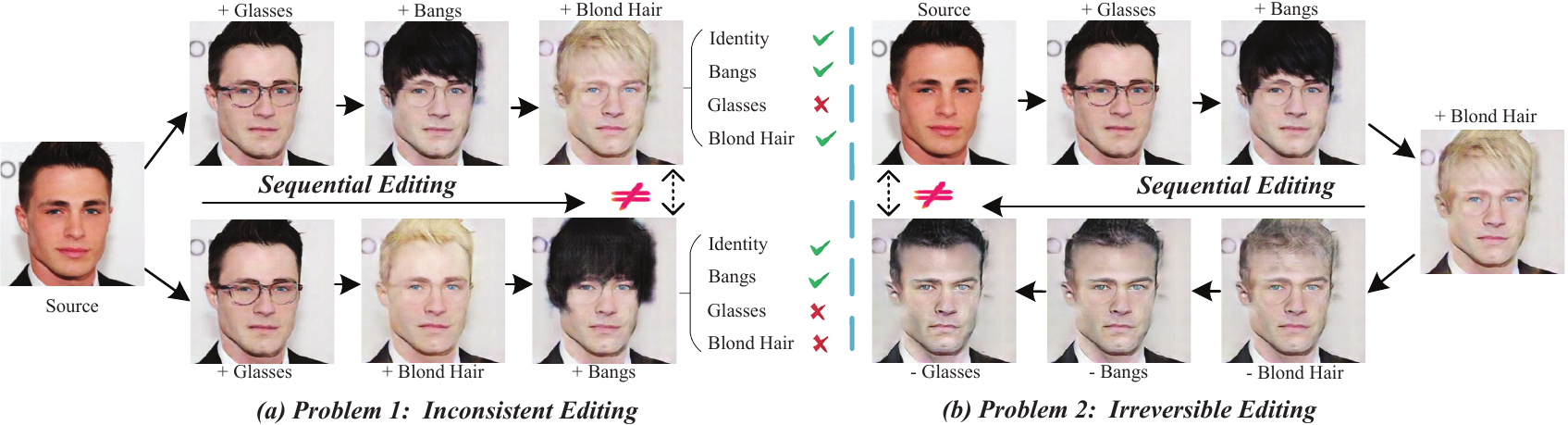}
    \caption{A case of inconsistent and irreversible editing.}
    \label{fig:problem}

\end{figure*}

1) We define three concepts for facial image editing, i.e., continuity, consistency, and reversibility. Then, we propose a novel model named CCR to achieve the desired goal;

2) We propose a progressive training strategy to train CCR, and then propose a criterion that can be used to determine whether a model is continuous, consistent, and reversible;

3) We draw an important conclusion that the necessary conditions for editing continuity, consistency, and reversibility are: a generalized encoder, disentangled style codes, and accurate attention regions.

\section{Proposed method}
In this section, we 1) describe the label structure and editing path; 2) present the framework of CCR model; 3) exhibit its training objective; 4) give a criterion to evaluate CCR model.
\subsection{Preliminary}
\begin{figure}[htb]
\centering
\includegraphics[width=0.48\textwidth]{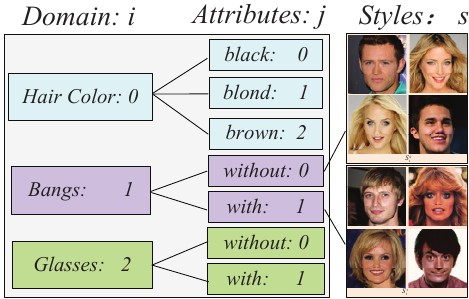}
\caption{ The introduced hierarchical tree structure \cite{li2021} with multi-domain and various attributes.}
\label{fig:label_org}
\end{figure}

We denote a facial image $x$ that has the attribute $j$ for the domain $i$ as $x_{i}^{j}$. In addition, $x_{i}^{j}$ has its style $s_{i}^{j}$. Assuming that the editing path is from hair color domain to bangs domain and then to glasses domain, we sample a facial image randomly, male/female with black hair, without bangs, and without glasses. To present a clear editing path, we use a fixed attribute editing order, as shown in Fig. \ref{fig:label_org}. Note that we can train CCR model in an arbitrary editing path. One editing path is:
\begin{equation}\label{eq1}
x_{0}^{0} \rightarrow x_{0}^{1} \rightarrow x_{1}^{1} \rightarrow x_{2}^{1}
\end{equation}
$x_{2}^{1}$ in Eq. (\ref{eq1}) should be with blond hair, bangs, and glasses. This is our defined concept of editing continuity. Another editing path is:
\begin{equation}\label{eq2}
x_{0}^{0} \rightarrow x_{0}^{1} \rightarrow x_{2}^{1} \rightarrow x_{1}^{1}
\end{equation}
$x_{1}^{1}$ in Eq. (\ref{eq2}) should be consistent with $x_{2}^{1}$ in Eq. (\ref{eq1}) even though swapping attribute editing orders. This is our defined concept of editing consistency. The reversible editing path is defined as:
\begin{equation}\label{eq3}
x^{\prime}{^{0}_{0}}\leftarrow x^{\prime}{^{1}_{0}} \leftarrow x^{\prime}{^{1}_{1}} \leftarrow x^{1}_{2}
\end{equation}
$x^{\prime}{^{0}_{0}}$ should be a face with black hair, without bangs, and without glasses, and $x^{\prime}{^{0}_{0}}$ should be equal to $x_{0}^{0}$.

\subsection{Framework}
\label{framework}
\begin{figure*}[!ht]
\centering
\includegraphics[width=\textwidth]{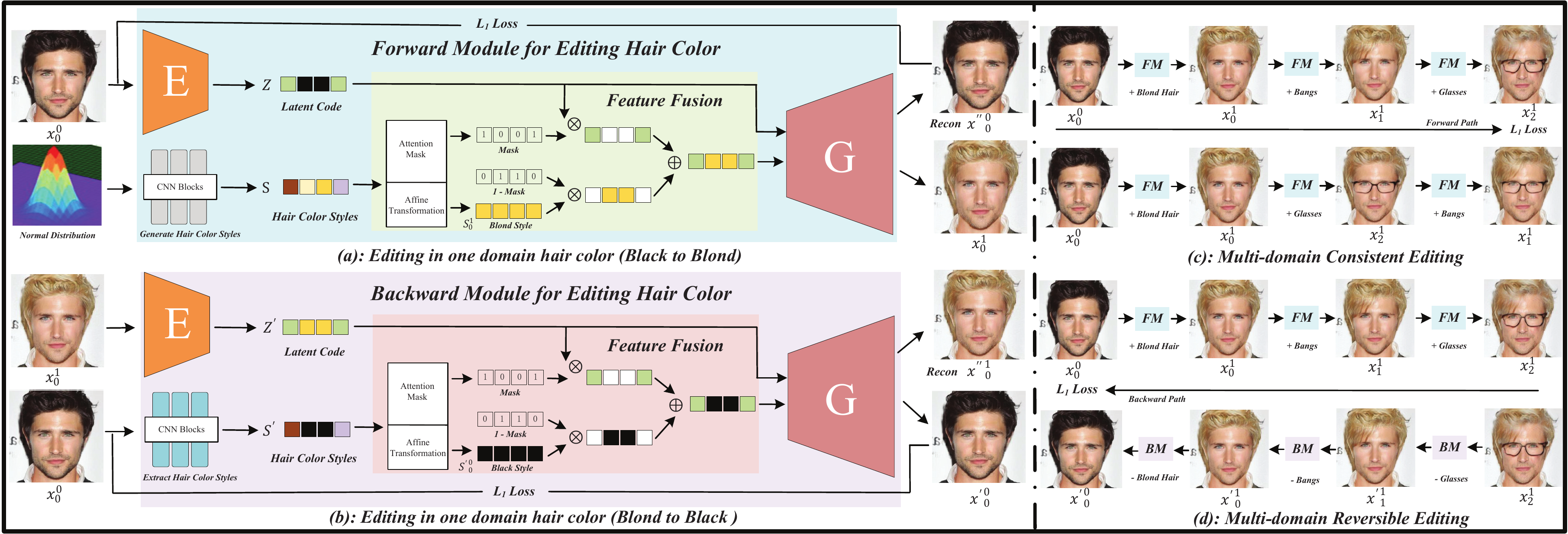}
\caption{The training pipeline of our method, which includes: a) Forward Module for editing in one domain; b) Backward Module for editing in one domain; c) Multi-domain Consistent Editing; d) Multi-domain Reversible Editing. Notice that the discriminator $D$ is not shown but is used for training.}
\label{fig:framework}
\end{figure*}

The framework of CCR model is shown in Fig. \ref{fig:framework}. We should train CCR model that can edit a facial image in a specific domain firstly, e.g., from black hair to blond hair in the domain of hair color in Fig. \ref{fig:framework}(a). The encoder $E$ aims to encode a facial image to a latent code $z = E(x_{0}^{0})$. $z$ contains hair color and facial identity information, etc. The CNN blocks learn to generate hair color styles $s$ from a normal distribution \cite{Shanmugam1993}. $z$ and $s$ have the same dimension. We introduce an attention module $M$ \cite{Moore2001a,Pumarola2018a} and an affine transformation $T$ to focus on specific regions and disentangles hair color styles respectively, i.e., $m=M(s)$, $t=T(s)$. Then, we fuse $m$, $z$, and $t$ using $\sigma(m) \cdot z+(1-\sigma(m)) \cdot t$, where $\sigma(\cdot)$ is the sigmoid function, and $\sigma(m)$ is an attention mask. As noted in \cite{li2021}, this design can avoid global manipulations like background and illumination during translations with little additional calculation and no regularization objective. Finally, the fused feature is fed to $G$ and generated an edited facial image $x_{0}^{1}$ with blond hair. $L_{1}$ is a loss function \cite{Bot2010,Mescheder2018a} to measure the spatial distance between $x_{0}^{0}$ and $x_{0}^{\prime \prime 0}$.

Fig. \ref{fig:framework}(b) illustrates the editing pipeline from blond hair to black hair. Different from forward editing process, CCR receives source facial image $x_{0}^{0}$ and edited image $x_{0}^{1}$ as inputs. The CNN blocks in the backward module extract hair color styles from source facial image. We expect to return a facial image $x_{0}^{\prime 0}$ with black hair that is equal to $x_{0}^{0}$. We simplify the forward module and the backward module as $FM$ and $BM$. Fig. \ref{fig:framework}(c) exhibits a case of consistent editing in multi-domain. Its corresponding editing path is Eqs. (\ref{eq1}) and (\ref{eq2}). Fig. \ref{fig:framework}(d) describes the case of reversible editing in multi-domain. Its corresponding editing is Eqs. (\ref{eq1}) and (\ref{eq3}). We can easily observe that editing continuity is an essential premise to achieve the goal of editing consistency and reversibility.

\textbf{Training strategy}. Note that the translated modules ($FM$ and $BM$) for a different domain are independent while the encoder ($E$) and generator ($G$) are shared and general. The shared part is regarded as an interface that matches multiple domains or categories. We use a progressive way to train our model. It contains three parts: 1) using an end-to-end strategy to train the shared encoder and generator. 2) using a one-to-one strategy to train different domains translated modules based on the well-trained shared parts. We can obtain independent translated modules for different domains. 3) selecting arbitrary editing paths and finetuning all the well-trained modules. Only updating the parameters of selected translated modules while freezing the parameters of shared and other translated modules in each iteration. Therefore, we can select and test any translated modules independently for translation to realize sequential editing.

\subsection{Optimization objective}
\textbf{Adversarial loss}. The model should treat all the generated facial images as fakes \cite{wei2018reconstructible,li2019discriminative,Miyato2018a}, except for the source image. The discriminator $D$ is used to evaluate whether a facial image is real or not. The adversarial objective is defined as:
\begin{equation}
\begin{aligned}
&\mathcal{L}_{adv}=\mathbb{E}\left[\log \left(D\left(x_{0}^{0}\right)\right)\right]+\mathbb{E}\left[\log \left(D\left(1-x_{0}^{1}\right)\right)\right]\\
&+\mathbb{E}\left[\log \left(D\left(1-x_{1}^{1}\right)\right)\right]+\mathbb{E}\left[\log \left(D\left(1-x_{2}^{1}\right)\right)\right]\\
&+\mathbb{E}\left[\log \left(D\left(1-x_{1}^{\prime 1}\right)\right)\right]+\mathbb{E}\left[\log \left(D\left(1-x_{0}^{\prime 1}\right)\right)\right]\\
&+\mathbb{E}\left[\log \left(D\left(1-x_{0}^{\prime 0}\right)\right)\right]
\end{aligned}
\label{eq:advloss}
\end{equation}
where, $x_{0}^{0}$ is a source facial image. $x_{0}^{1}$, $x_{1}^{1}$ and $x_{2}^{1}$ are the translated images using $FM$, while $x_{1}^{\prime 1}$, $x_{0}^{\prime 1}$  and $x_{0}^{\prime 0}$ are the reversed facial images using $BM$. ${\mathcal{L}}_{adv}$ encourages the generator $G$ to yield a high-quality facial image to support editing continuity.

\textbf{Reconstruction loss}. The model should reconstruct source image $x_{0}^{0}$ in the forward editing process. To achieve editing continuity, the model should also reconstruct an edited facial image $x_{0}^{1}$ in the backward editing process. Thus, we introduce a reconstructive objective \cite{Zhao2018a} to force CCR to reduce the difference between source image and its corresponding edited one. We have the following reconstruction loss:
\begin{equation}
\begin{aligned}
\mathcal{L}_{r e c}&=\mathbb{E}\left[\left\|x_{0}^{0}-x_{0}^{\prime \prime 0}\right\|_{1}\right]+\mathbb{E}\left[\left\|x_{0}^{1}-x_{0}^{\prime \prime 1}\right\|_{1}\right] \\
&+\mathbb{E}\left[\left\|x_{1}^{1}-x_{1}^{\prime \prime 1}\right\|_{1}\right]+\mathbb{E}\left[\left\|x_{2}^{1}-x_{2}^{\prime \prime 1}\right\|_{1}\right]
\end{aligned}
\label{eq:recloss}
\end{equation}
$\mathcal{L}_{r e c}$: 1) encourages $E$ to retain global facial features; 2) promotes $G$ to generate a facial image that is equal to the source one; and 3) forces CNN blocks to extract attribute styles from a given facial image accurately.

\textbf{Consistency loss}. To retain editing consistency, we introduce the following consistency loss:
\begin{equation}
\mathcal{L}_{con}=\mathbb{E}\left[\left\|x_{1}^{1}-x_{2}^{1}\right\|_{1}\right]
\label{eq:conloss}
\end{equation}
$\mathcal{L}_{con}$ aims to reduce the difference between $x_{1}^{1}$ and $x_{2}^{1}$. The two editing results are expected to have the same attributes and facial identity.

\textbf{Reversibility loss}. To achieve editing reversibility, we introduce the following reversibility loss:
\begin{equation}
\begin{aligned}
\mathcal{L}_{rev}&=\mathbb{E}\left[\left\|x_{1}^{1}-x_{1}^{\prime 1}\right\|_{1}\right]\\
&+\mathbb{E}\left[\left\|x_{0}^{1}-x_{0}^{\prime 1}\right\|_{1}\right]+\mathbb{E}\left[\left\|x_{0}^{0}-{x^{\prime}}_{0}^{0}\right\|_{1}\right]
\end{aligned}
\label{eq:revloss}
\end{equation}
$\mathcal{L}_{rev}$ encourages the model to return a facial image that is equal to the source one at every stage.

\textbf{Style loss}. The extracted style codes for the edited facial image in $BM$ should be equal to the generated style code from $FM$ \cite{Huang2018a,wei2018reconstructible,Wang2019a,Choi2018}. We denote the function of extract style codes as $F(\cdot)$, then we have the following style loss:
\begin{equation}
\begin{aligned}
\mathcal{L}_{sty}&=\mathbb{E}\left[\left\|F\left(x_{0}^{1}\right)-s_{0}^{1}\right\|_{1}\right]\\
&+\mathbb{E}\left[\left\|F\left(x_{1}^{1}\right)-s_{1}^{1}\right\|_{1}\right]+\mathbb{E}\left[\left\|F\left(x_{2}^{1}\right)-s_{2}^{1}\right\|_{1}\right]
\end{aligned}
\label{eq:styloss}
\end{equation}
$\mathcal{L}_{sty}$ keeps the consistency between the generated and extracted styles. The objective is to ensure that we can return a facial image with the same attributes.

\textbf{Overall loss}. We optimize the model in an editing path within the following objective:
\begin{equation}
\begin{aligned}
\min _{E, G, F}\max _{D} \mathcal{L}_{a d v}&+\lambda_{r e c} \mathcal{L}_{r e c}\\
&+\lambda_{\text {con }} \mathcal{L}_{\text {con }}+\lambda_{\text {rev }} \mathcal{L}_{r e v}+\lambda_{\text {sty }} \mathcal{L}_{s t y}
\end{aligned}
\label{eq:totalloss}
\end{equation}
where $\lambda_{r e c}$, $\lambda_{con}$, $\lambda_{rev}$ and $\lambda_{sty}$ are positive hyper-parameters that control reconstruction, consistency, reversibility, and style objectives. The full objective guarantees CCR is a continuous, consistent, and reversible model.

\subsection{Criterion}
To evaluate continuity and consistency, we must judge all the attributes are added to the final desired editing facial image. For this purpose, we define the editing accuracy in multi-domain as:
\begin{equation}
EAC=\frac{\sum_{1}^{m} \sum_{i}^{n} H\left(cls\left(x_{i}\right), y_{i}\right)}{m*n}
\label{eq:eac}
\end{equation}
where, $m$ denotes the number of edited facial images and $n$ indicates the number of domains. $x_{i}$ denotes an edited facial image in domain $i$, while $y_{i}$ denotes its corresponding true label. $cls$ is a well-trained classifier, which outputs the predicted binary label for a facial image. $H$ denotes Hamming distance \cite{Norouzi2012}, which computes the number of different bits in binary. EAC computes the sum of editing accuracy for each domain.

To evaluate reversibility, we need to determine if all the edited attributes are removed from the final editing facial image, i.e., remove accuracy:
\begin{equation}
RAC=1-\frac{\sum_{1}^{m} \sum_{i}^{n} H\left(cls\left(x^{\prime}_{i}\right), y_{i}\right)}{m*n}
\label{eq:rac}
\end{equation}
where $x^{\prime}_{i}$ denotes a reversed facial image in domain $i$.

Evaluating a model with EAC and RAC only from attribute level is not sufficient. We further assess the edited facial image quantificationally by using multi-scale metrics, e.g., MSE \cite{Guo2004a}, RMSE \cite{Chai2014a}, PSNR \cite{Hore2010a}, UQI \cite{Wang2002}, SSIM \cite{Wang2004}, MS-SSIM \cite{Snell2017}, VSI\cite{zhang2014vsi}, VIF\cite{sheikh2006image}, FSIM\cite{zhang2011fsim}, GMSD\cite{xue2013gradient}, LPIPS\cite{zhang2018unreasonable}, DISTS\cite{ding2020image}. The model has higher values in all the above metrics, and the better it is. We evaluate CCR from both attribute and image levels, and the results are more reliable than that with a single metric.

\section{Experimental results and analysis}
In this section, we 1) describe the experimental implementation details; 2) present single attribute editing results; 3) show the editing results of continuity and consistency; 4) illustrate the editing results of reversibility; and 5) provide the results of ablation studies.
\subsection{Implementation details}
We train CCR model on the CelebA-HQ \cite{Karras2017} dataset, which contains 30000 facial images with label annotations \cite{Liu2015}. Five attributes are chosen in three domains, i.e., hair color, bangs, glasses. We divide the datasets into training and testing sets containing 27000 and 3000 images, respectively. Our experimental environment is based on Lenovo Intelligent Computing Orchestration (LiCO), a software solution that simplifies the use of clustered computing resources for artificial intelligence (AI) model development and training. All experiments are conducted in PyTorch 1.7 \cite{Paszke2017a}, CUDA 10.2, and CUDNN 10.2 with 8 NVidia Tesla V100 (32G) dual-channel graphics processing units (GPUs).

\begin{figure}[!ht]
\centering
\includegraphics[width=0.45\textwidth]{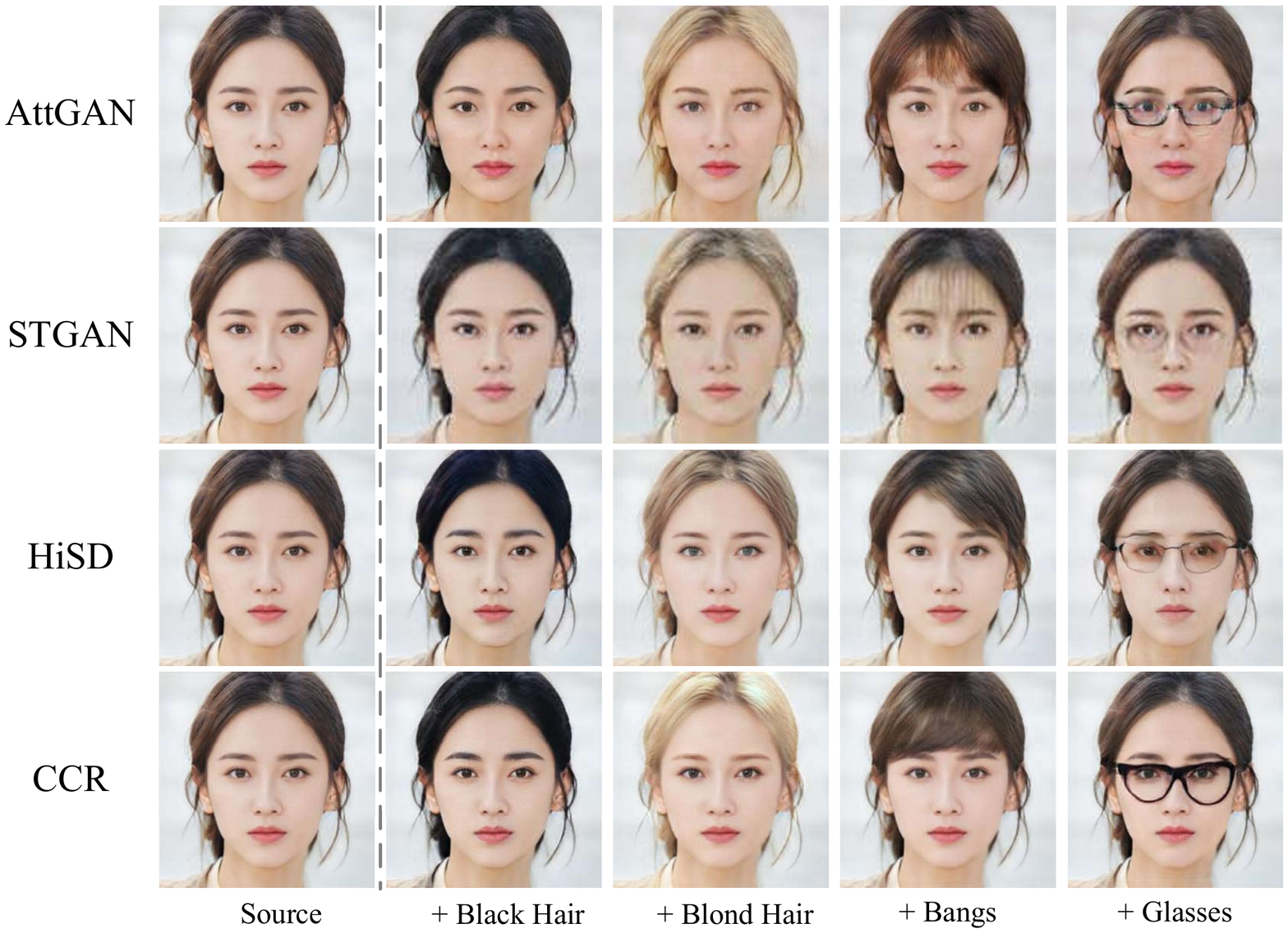}
\caption{ The compared editing results with AttGAN \cite{He2019}, STGAN \cite{Liu2019}, HiSD \cite{li2021} in one domain.}
\label{fig:single_attribute}
\end{figure}

\subsection{Single Attribute Editing}
We evaluate the performance of CCR in one domain and compare it with three competing methods, i.e., AttGAN \cite{He2019}, STGAN \cite{Liu2019}, and HiSD \cite{li2021}. AttGAN and STGAN are designed for facial image editing. HiSD is the current challenging facial image editing method. The qualitative results are shown in Fig. \ref{fig:single_attribute}. We have three observations: 1) AttGAN and STGAN are still limited to editing glasses; 2) AttGAN attempts to change the color of bangs while STGAN has bangs that are not obvious; 3) HiSD yields a good editing result compared to AttGAN and STGAN. Unfortunately, the hair is not edited to blond completely. In comparison, CCR can edit the desired attributes effectively and correctly and yield results with high image quality.

\begin{figure*}[t]
\centering
\includegraphics[width=\textwidth]{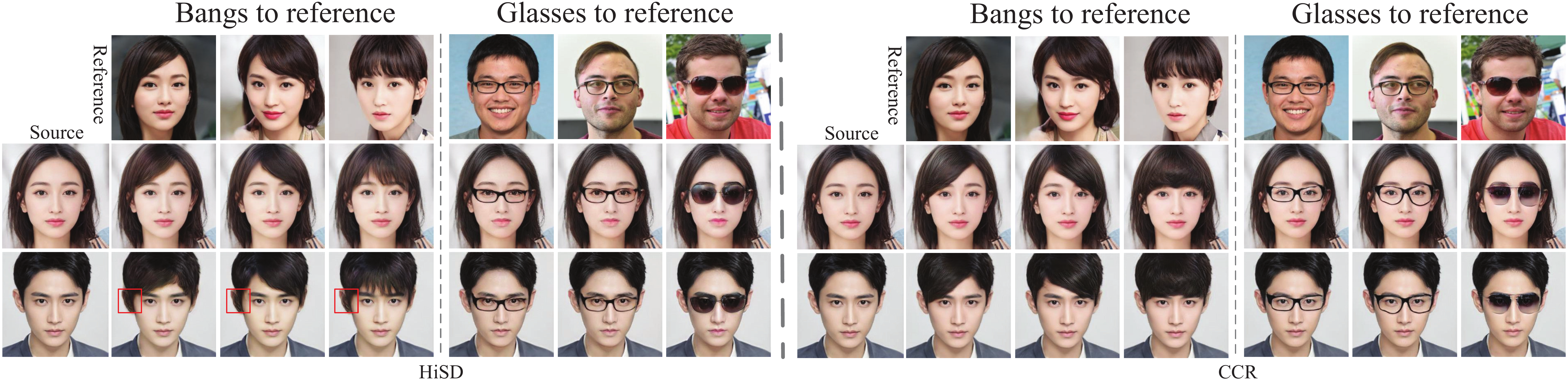}
\caption{The compared editing results with HiSD \cite{li2021}, where the styles are extracted from the reference facial images. Image artifacts are labeled with a red box in poorly edited images.}
\label{fig:reference}
\end{figure*}

\begin{table}[!ht]
  \caption{The compared single EAC with AttGAN \cite{He2019}, STGAN \cite{Liu2019}, HiSD \cite{li2021} in one domain.}
  \centering
  \resizebox{\linewidth}{!}{
  \begin{tabular}{|c|c|c|c|c|c|c|}
        \hline
        %\cline{1-1}
        Configuration    &Bangs	&Black Hair	&Blond Hair	&Brown Hair	&Glasses\\\hline
        AttGAN \cite{He2019}   &0.8640 	&0.9237 	&0.7225 	&0.5354 	&0.9437    \\\hline
        STGAN  \cite{Liu2019}   &0.7113 	&0.8274 	&0.4158 	&0.5229	&0.9111    \\\hline
        HiSD  \cite{li2021}    &0.7406 	&0.8669 	&0.6864 	&0.5768 	&0.7769     \\\hline
        CCR       &0.9066 	&0.8935 	&0.8064 	&0.6100 	&0.9265    \\\hline
        \end{tabular}}

  \label{tab:accuracy}
\end{table}

\begin{figure*}[!t]
\centering
\includegraphics[width=\textwidth]{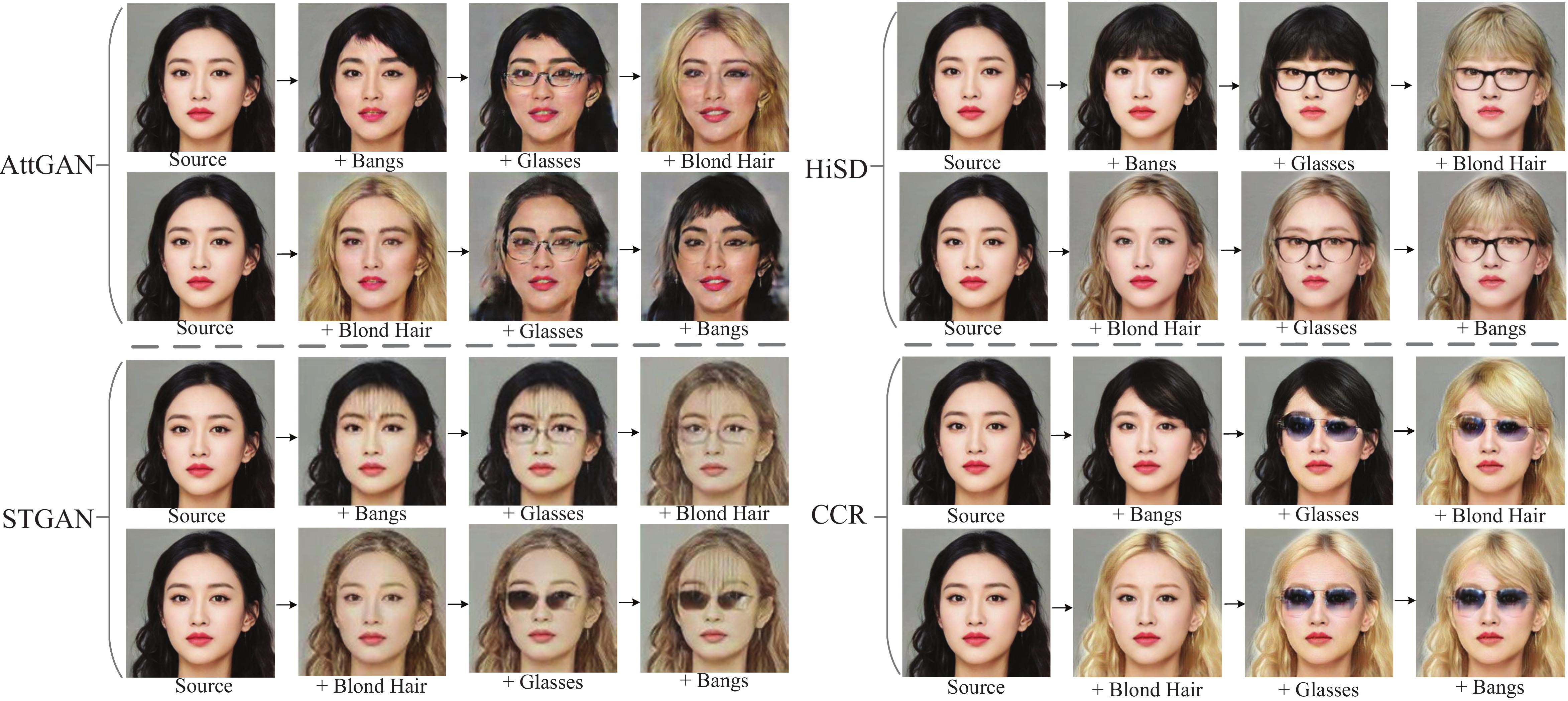}
\caption{The editing result of the editing continuity and consistency compared to AttGAN \cite{He2019}, STGAN \cite{Liu2019}, HiSD \cite{li2021}.}
\label{fig:consistency}
\end{figure*}

\begin{table*}[!ht]
\caption{The quantitative results of evaluating editing continuity and consistency using EAC and image quality metrics. $\uparrow$ and $\downarrow$ denote the higher and the lower the better, respectively.Methods with the best and runner-up performances are colored with {\textcolor{red}{red}} and {\textcolor{blue}{blue}}, respectively }
\centering
\resizebox{0.7\textwidth}{1.2cm}{%
\begin{tabular}{|c|c|c|c|c|c|c|c|c|}
\hline
\multirow{2}{*}{Method} &
  \multicolumn{2}{c|}{EAC $\uparrow$} &
  \multirow{2}{*}{MSE $\downarrow$} &
  \multirow{2}{*}{RMSE $\downarrow$} &
  \multirow{2}{*}{UQI $\uparrow$} &
  \multirow{2}{*}{SSIM $\uparrow$} &
  \multirow{2}{*}{PSNR $\uparrow$} &
  \multirow{2}{*}{MS-SSIM $\uparrow$} \\ \cline{2-3}
       & Path 1   & Path 2   &            &            &            &            &                         &            \\ \hline
AttGAN \cite{He2019} & 36.46\% & 31.15\% & 0.0272 & 0.1569 & 0.8771 & 0.8261 & 16.5303  & 0.8330 \\ \hline
STGAN \cite{Liu2019} & 40.24\% & 47.53\% & 0.0115 & 0.1003 & 0.8610 & \textcolor[rgb]{0.00,0.07,1.00}{0.8788} & \textcolor[rgb]{0.00,0.07,1.00}{20.5773}  & \textcolor[rgb]{0.00,0.07,1.00}{0.9109} \\ \hline
HiSD \cite{li2021}  & \textcolor[rgb]{0.00,0.07,1.00}{52.99\%} & \textcolor[rgb]{0.00,0.07,1.00}{57.19\%} & \textcolor[rgb]{0.00,0.07,1.00}{0.0107} & \textcolor[rgb]{0.00,0.07,1.00}{0.0997} & \textcolor[rgb]{0.00,0.07,1.00}{0.9023} & 0.8625 & 20.3696  & 0.8989 \\ \hline
CCR    & \textcolor[rgb]{1.00,0.00,0.00}{59.41\%} & \textcolor[rgb]{1.00,0.00,0.00}{60.49\%} & \textcolor[rgb]{1.00,0.00,0.00}{0.0096} & \textcolor[rgb]{1.00,0.00,0.00}{0.0943} & \textcolor[rgb]{1.00,0.00,0.00}{0.9173} & \textcolor[rgb]{1.00,0.00,0.00}{0.8847} & \textcolor[rgb]{1.00,0.00,0.00}{20.8347}  & \textcolor[rgb]{1.00,0.00,0.00}{0.9196} \\ \hline
\end{tabular}%
}

\label{tab:cc}
\end{table*}

\begin{figure*}[!t]
\centering
\includegraphics[width=\textwidth]{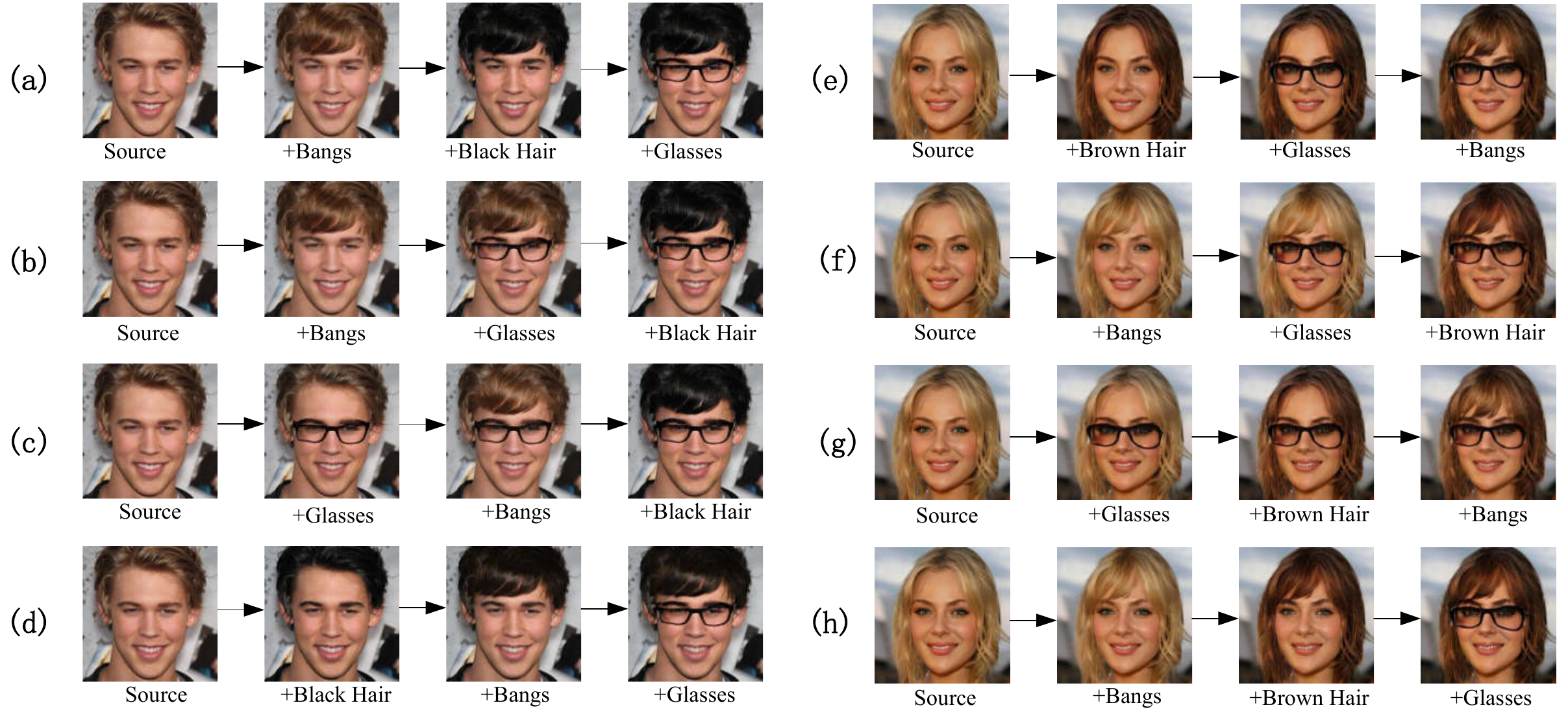}
\caption{The editing results of the editing continuity and consistency by CCR. The editing paths for (a) source$\to$+bangs$\to$+black hair$\to$+glasses; (b) source$\to$+bangs$\to$+glasses$\to$+black hair; (c) source$\to$+glasses$\to$+bangs$\to$+black hair; (d) source$\to$+black hair$\to$+bangs$\to$+glasses; (e) source-$\to$+brown hair$\to$+ glasses$\to$+bangs; (f) source$\to$+bangs$\to$+glasses$\to$+brown hair; (g) source$\to$+glasses$\to$+brown hair$\to$+bangs and (h) source$\to$+bangs$\to$+brown hair$\to$+glasses.}
\label{fig:consistency_our}
\end{figure*}

The proposed CCR can extract style codes from different reference images. These codes can be used to perform custom facial image editing. The qualitative results in comparison to HiSD are shown in From Fig. \ref{fig:reference}. We can observe: 1) both methods can extract diverse codes from reference images and add corresponding styles to source facial images, i.e., bangs and glasses; 2) the styles of bangs extracted using HiSD are maybe over reliant their reference images. This reliance sometimes results in image artifacts, as shown in the red box in Fig. \ref{fig:reference}. CCR uses reconstruction loss in multi-domain and will not yield such cases. The experimental results show that CCR can extract style codes from reference images accurately and perform custom facial image editing well.

We evaluate the attribute editing accuracy quantitatively by re-using a pre-trained classification model, which can attain an accuracy of 94.5\% on the test set \cite{Liu2019}. The experimental results are shown in Table \ref{tab:accuracy}. CCR outperforms HiSD on all the attributes. Besides, it exceeds the professional editing methods of AttGAN and STGAN on most of attributes, e.g., bangs, blond and brown hair. We are slightly less accurate than AttGAN in editing glasses. However, we observe that AttGAN has a poor image quality, which results in a woman who do not wear glasses being misjudged as wearing glasses, as shown in Fig. \ref{fig:single_attribute}. CCR can also add diverse glasses styles to a person in comparison to AttGAN and STGAN.

\subsection{Continuity and Consistency}
As noted in Sec. \ref{framework}, editing continuity is an essential premise to consistency. Thus, we present these two concepts into one subsection. The qualitative results in comparison to three competing methods are shown in Fig. \ref{fig:consistency}.

\textbf{Editing continuity}: AttGAN and STGAN can not retain the edited attributes. The image quality deteriorates, and facial identity is lost in performing editing sequentially. The results illustrate that both methods do not have the property of editing continuity. HiSD can retain the edited attributes while preserving facial identity. However, the blond hair is disturbed by black hair and not edited successfully. CCR can retain the edited attributes and preserve facial identity well. Furthermore, the blond is edited completely and not disturbed by black hair. The experimental results show that CCR fully considers editing continuity, i.e., users can re-edit an edited facial image while preserving facial identity and retaining all the previously edited attributes unchanged.

\textbf{Editing consistency}: The final editing results of AttGAN and STGAN have a larger margin when swapping attribute editing orders. The reason is that both methods are limited to editing continuity and can not be realized editing consistency. Although HiSD can retain the edited attributes, it has an inconsistent editing result when swapping attribute editing orders, e.g., swap the style codes of bangs and glasses. It is owing to that HiSD does not consider editing consistency. CCR can obtain a consistent and high-quality editing result in two different editing paths. The successful editing results are attributed to adversarial, reconstruction, and consistency loss in multi-domain. The experimental results show that CCR fully considers editing continuity and consistency. Users can generate high-quality editing results while retaining their desired attributes after sequential editing. CCR allows users to produce a consistent editing result despite swapping attribute editing orders.

\begin{figure*}[!t]
\centering
\includegraphics[width=\textwidth]{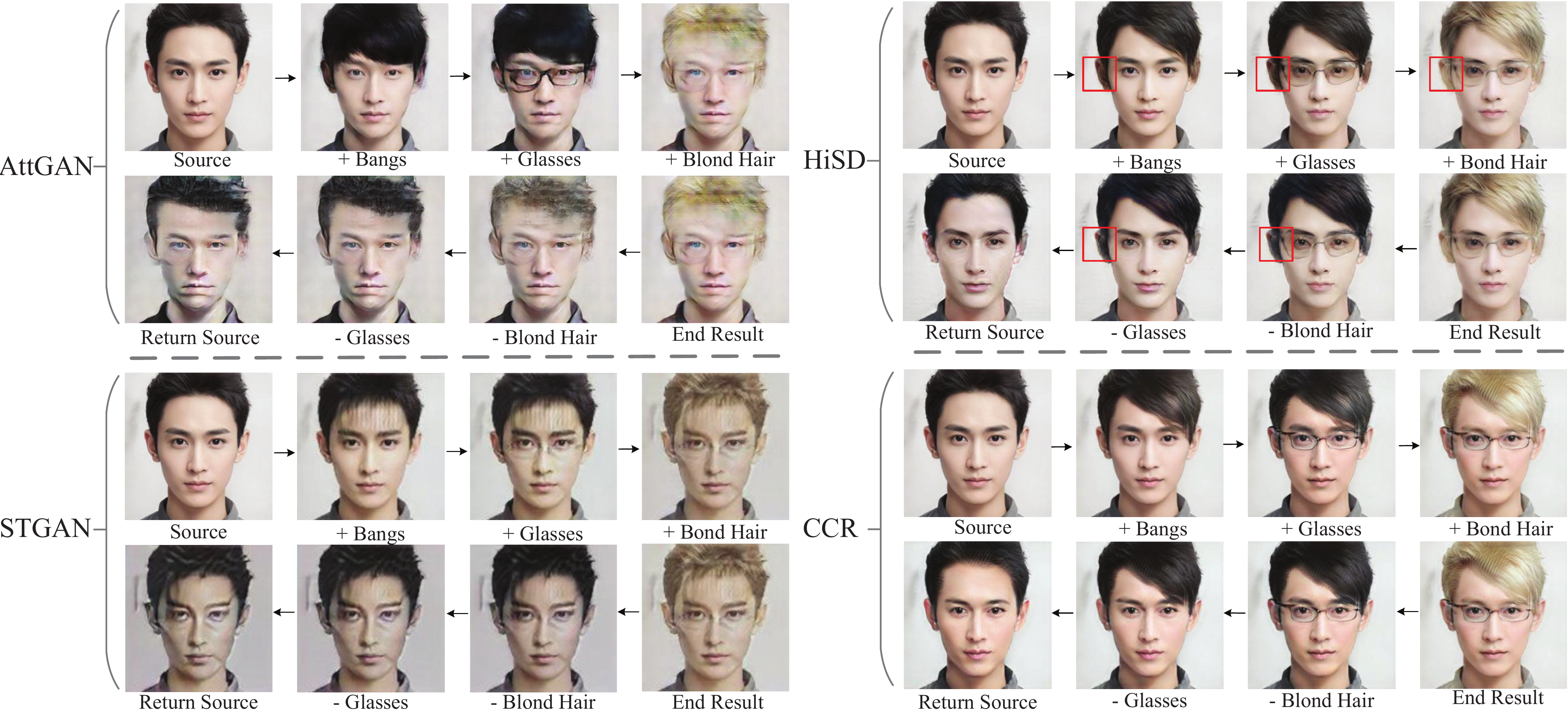}
\caption{The editing result of the editing reversibility compared to AttGAN \cite{He2019}, STGAN \cite{Liu2019}, HiSD \cite{li2021}. Image artifacts are labeled with a red box in poorly edited images.}
\label{fig:reversibility}
\end{figure*}

We quantitatively analyze the results of two different editing paths, as shown in Table \ref{tab:cc}. For each method, the above path is denoted as Path 1, while the above path is denoted as Path 2. We can see that CCR has a higher editing accuracy on both editing paths than all the competing methods. Our method is better than peer methods in all the image quality evaluation metrics. STGAN has a lower editing accuracy than HiSD, while it has a better performance in image quality evaluation metrics. We visualize the editing results of STGAN and find that the output image is too smooth to retain its raw features, as shown in Fig. \ref{fig:consistency}. The processing of pixels in STGAN makes a better performance in image quality evaluation metrics, while its editing accuracy is lower than HiSD. In Table \ref{tab:cc}, the quantitative experimental results show that CCR performs well in both editing accuracy and image quality evaluation metrics. It is due to that CCR fully considers editing continuity and consistency. More editing results of editing continuity and consistency in Fig. \ref{fig:consistency_our} and extra evaluated results in Table \ref{tab:continuity} further demonstrate the well performance of CCR.

\begin{table*}[!ht]
\caption{The quantitative results for comparing editing continuity and consistency on AttGAN \cite{He2019}, STGAN \cite{Liu2019} HiSD \cite{li2021} and CCR based more image quality metrics.}
\centering
\resizebox{0.6\textwidth}{1.2cm}{
\begin{tabular}{|c|c|c|c|c|c|c|c|}
\hline
\multirow{2}{*}{Method} &
  \multirow{2}{*}{VSI $\uparrow$} &
  \multirow{2}{*}{VIF $\uparrow$} &
  \multirow{2}{*}{FSIM $\uparrow$} &
  \multirow{2}{*}{GMSD $\uparrow$} &
  \multirow{2}{*}{LPIPS $\uparrow$} &
  \multirow{2}{*}{DISTS $\uparrow$} \\
       &        &            &            &            &            &       \\ \hline
AttGAN  & 0.9620 & 0.3815 & 0.8837 & 0.8684 & 0.8658 & 0.8825 \\ \hline
STGAN  & 0.9624 & 0.3774 & 0.9037 & 0.8902 & 0.9340  & 0.8971 \\ \hline
HiSD  & \textcolor[rgb]{0.00,0.07,1.00}{0.9743} & \textcolor[rgb]{0.00,0.07,1.00}{0.4857} & \textcolor[rgb]{0.00,0.07,1.00}{0.9216} & \textcolor[rgb]{0.00,0.07,1.00}{0.9071} & \textcolor[rgb]{0.00,0.07,1.00}{0.9344} & \textcolor[rgb]{0.00,0.07,1.00}{0.9246} \\ \hline
CCR    &\textcolor[rgb]{1.00,0.00,0.00} {0.9761} & \textcolor[rgb]{1.00,0.00,0.00}{0.5554} & \textcolor[rgb]{1.00,0.00,0.00}{0.9338} & \textcolor[rgb]{1.00,0.00,0.00}{0.9177} & \textcolor[rgb]{1.00,0.00,0.00}{0.9385} & \textcolor[rgb]{1.00,0.00,0.00}{0.9251} \\ \hline
\end{tabular}}
\label{tab:continuity}
\end{table*}

\subsection{Reversibility}
\begin{figure*}[!t]
\centering
\includegraphics[width=\textwidth]{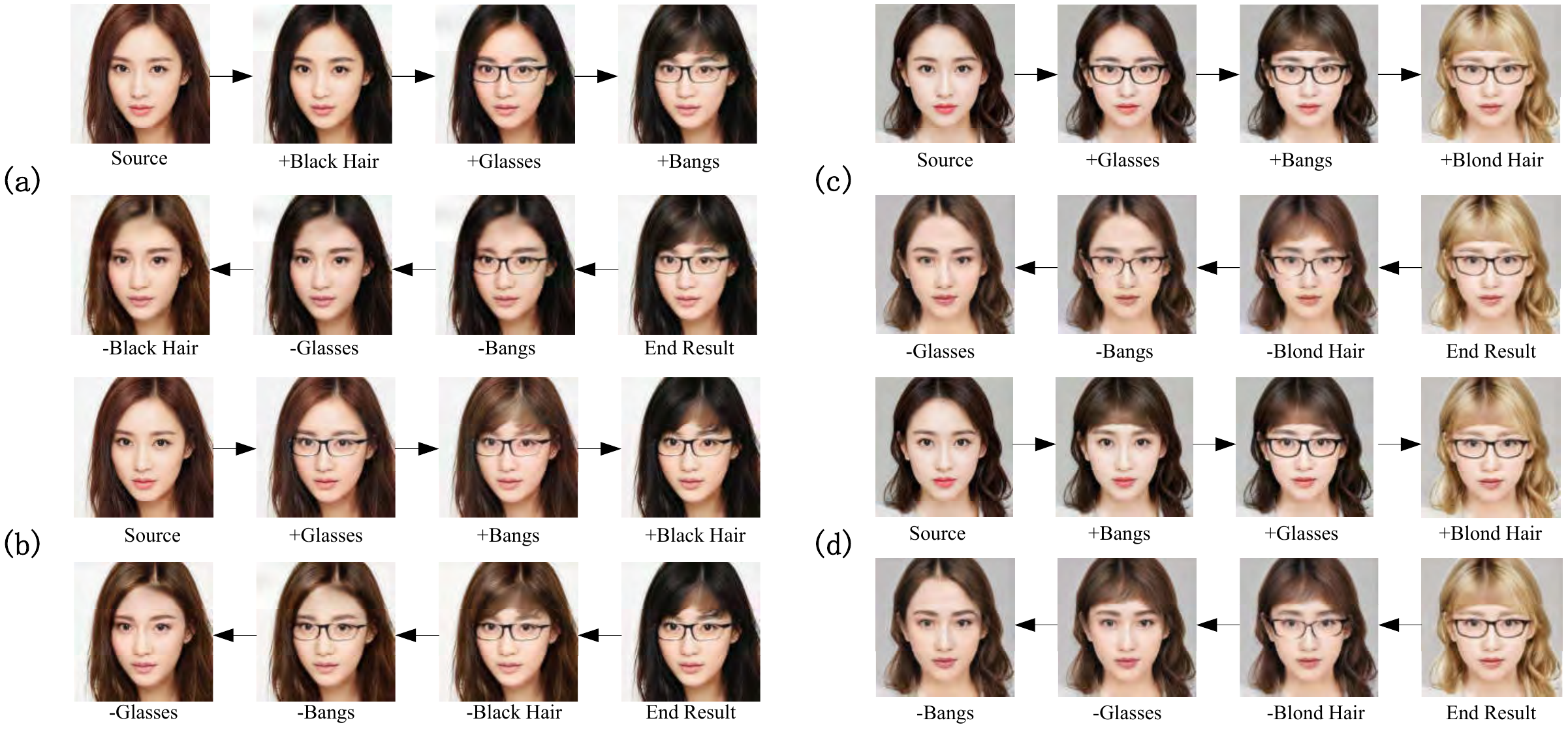}
\caption{The editing results of the editing reversibility by CCR. The editing paths for (a) source$\to$+black hair$\to$+glasses$\to$+bangs$\to$-bangs$\to$-glasses$\to$-black hair; (b) source$\to$+glasses$\to$+bangs$\to$+black hair$\to$-black hair$\to$-bangs$\to$-glasses; (c) source$\to$+glasses$\to$+bangs$\to$+blond hair$\to$-blond hair$\to$-bangs$\to$-glasses and (d) source$\to$+bangs$\to$+glasses$\to$+blond hair$\to$-blond hair$\to$-glasses$\to$-bangs.}
\label{fig:reversibility_our}
\end{figure*}

We further investigate the editing reversibility of CCR. We use RAC in Eq. \ref{eq:rac} to evaluate the attribute remove accuracy of the reversed facial image. The image quality metrics are used to evaluate the similarity between the source image and reversed one. The qualitative experimental results are shown in Fig. \ref{fig:reversibility}. AttGAN and STGAN can not add corresponding attributes to a person. Both methods can not generate high-quality reversed facial images and preserve facial identity. The results indicate that their editing is irreversible. HiSD can not edit blond hair completely since it suffers from attribute entanglement. Besides, it produces an image artifact when editing bangs, as shown in the red box in Fig. \ref{fig:reversibility}. These image artifacts remain when editing a facial image along an inverse editing path, returning a low-quality facial image and losing facial identity. The results indicate that HiSD does not consider editing reversibility. CCR can restore a source facial image well and remove the edited attributes correctly. The reason for the phenomenon is that CCR fully considers editing reversibility and multi-domain image reconstruction. The experimental results demonstrate that CCR is reversible and can achieve the goal of editing reversibility in comparison to the competing methods. The quantitative experimental results are shown in Table \ref{tab:reversibility}. AttGAN and STGAN have a high RAC value. However, both methods do not add desired attributes to a person but return a facial image that is similar to the source one. HiSD has a higher RAC value than AttGAN and STGAN, and it is superior to both methods on the image evaluation metrics. CCR outperforms the three competing methods in all the metrics. CCR can return a facial image that is closer to the source image than the three competing methods. It also can retain the quality of the reversed image and preserve facial identity. The quantitative experimental results show that CCR can well do facial image editing with reversibility. More reversible editing results in Fig. \ref{fig:reversibility_our} and quantitative results in Table \ref{tab:accurc} show that CCR can achieve editing reversibility in comparison to peer works.

\begin{table*}[!ht]
\caption{The quantitative results of evaluating editing reversibility using RAC and image quality metrics.}
\centering
\resizebox{0.65\textwidth}{1.2cm}{%
\begin{tabular}{|c|c|l|c|c|c|c|c|c|}
\hline
\multirow{2}{*}{Method} &
  \multicolumn{2}{c|}{\multirow{2}{*}{RAC $\uparrow$}} &
  \multirow{2}{*}{MSE $\downarrow$} &
  \multirow{2}{*}{RMSE $\downarrow$} &
  \multirow{2}{*}{UQI $\uparrow$} &
  \multirow{2}{*}{SSIM $\uparrow$} &
  \multirow{2}{*}{PSNR $\uparrow$} &
  \multirow{2}{*}{MS-SIM $\uparrow$} \\
       & \multicolumn{2}{c|}{}        &            &            &            &            &                         &            \\ \hline
AttGAN \cite{He2019} & \multicolumn{2}{c|}{86.60\%} & 0.0940 & 0.2986 & 0.7132 & 0.5926 & 10.7178  & 0.6513 \\ \hline
STGAN \cite{Liu2019}  & \multicolumn{2}{c|}{90.94\%} & 0.0843 & 0.2869 & 0.6772 & 0.5638 & 10.9477  & 0.7307 \\ \hline
HiSD \cite{li2021}  & \multicolumn{2}{c|}{\textcolor[rgb]{0.00,0.07,1.00}{94.48\%}} & \textcolor[rgb]{0.00,0.07,1.00}{0.0271} & \textcolor[rgb]{0.00,0.07,1.00}{0.1606} & \textcolor[rgb]{0.00,0.07,1.00}{0.8424} & \textcolor[rgb]{0.00,0.07,1.00}{0.7237} & \textcolor[rgb]{0.00,0.07,1.00}{16.0690}  & \textcolor[rgb]{0.00,0.07,1.00}{0.8153} \\ \hline
CCR    & \multicolumn{2}{c|}{\textcolor[rgb]{1.00,0.00,0.00}{94.97\%}} & \textcolor[rgb]{1.00,0.00,0.00}{0.0223} & \textcolor[rgb]{1.00,0.00,0.00}{0.1464} & \textcolor[rgb]{1.00,0.00,0.00}{0.8590} & \textcolor[rgb]{1.00,0.00,0.00}{0.7379} & \textcolor[rgb]{1.00,0.00,0.00}{16.8443}  & \textcolor[rgb]{1.00,0.00,0.00}{0.8366} \\ \hline
\end{tabular}%
}

\label{tab:reversibility}
\end{table*}

\begin{table*}[!ht]
\caption{The quantitative results for comparing editing reversibility on AttGAN \cite{He2019}, STGAN \cite{Liu2019} HiSD \cite{li2021} and CCR based more image quality metrics.}
\centering
\resizebox{0.6\textwidth}{1.2cm}{%
\begin{tabular}{|c|c|c|c|c|c|c|c|}
\hline
\multirow{2}{*}{Method} &
  \multirow{2}{*}{VSI $\uparrow$} &
  \multirow{2}{*}{VIF $\uparrow$} &
  \multirow{2}{*}{FSIM $\uparrow$} &
  \multirow{2}{*}{GMSD $\uparrow$} &
  \multirow{2}{*}{LPIPS $\uparrow$} &
  \multirow{2}{*}{DISTS $\uparrow$} \\
       &        &            &            &            &            &
\\ \hline
AttGAN  & 0.9249 & 0.1936 & 0.7782 & 0.8024 & 0.7235 & 0.7829 \\ \hline
STGAN  & 0.9260 & 0.2316 & 0.8176 & 0.8362 & \textcolor[rgb]{1.00,0.00,0.00}{0.8447} & 0.8100 \\ \hline
HiSD  & \textcolor[rgb]{0.00,0.07,1.00}{0.9473} & \textcolor[rgb]{0.00,0.07,1.00}{0.2987} & \textcolor[rgb]{0.00,0.07,1.00}{0.8450} & \textcolor[rgb]{0.00,0.07,1.00}{0.8549} & 0.8368 & \textcolor[rgb]{1.00,0.00,0.00}{0.8356}  \\ \hline
CCR    & \textcolor[rgb]{1.00,0.00,0.00}{0.9488} & \textcolor[rgb]{1.00,0.00,0.00}{0.3384} & \textcolor[rgb]{1.00,0.00,0.00}{0.8564} & \textcolor[rgb]{1.00,0.00,0.00}{0.86853} & \textcolor[rgb]{0.00,0.07,1.00}{0.8387} & \textcolor[rgb]{0.00,0.07,1.00}{0.8317}  \\ \hline
\end{tabular}%
}
\label{tab:accurc}
\end{table*}

\subsection{Ablation studies}
The encoder is expected to extract different image features rather than to focus on reconstructing facial identity, i.e., the encoder should be a generalized one. For this purpose, we add the identity loss \cite{Deng2019,Tan2021a} to encoder training process while leaving the rest of the training process unchanged, forcing CCR tends to reconstruct facial identity. The quantitative results of with/without identity loss on single attribute editing accuracy, multi-domain EAC, and RAC are shown in Table \ref{tab:ablation}. All the accuracy declines when adding identity loss to train the encoder except RAC. The results indicate that identity loss makes the encoder focus on facial identity while neglecting other significant image information. The encoder trained with a specific loss can not extract global features and pass them to the next editing process, resulting in a sharp decline in editing accuracy. After investigating the training process of encoder, we experimentally conclude that the generalized encoder is a necessary condition to achieve the goal of editing continuity, consistency, and reversibility.

\begin{table*}[!ht]
\caption{The quantitative results of with/without identity loss on single EAC, multi-domain EAC, and RAC.}
\centering
\resizebox{0.83\textwidth}{!}{%
\begin{tabular}{|c|c|c|c|c|c|c|c|c|}
\hline
\multirow{2}{*}{Method} &
  \multicolumn{2}{c|}{EAC $\uparrow$} &
  \multirow{2}{*}{RAC $\uparrow$} &
  \multirow{2}{*}{Bangs} &
  \multirow{2}{*}{Black Hair} &
  \multirow{2}{*}{Blond Hair} &
  \multirow{2}{*}{Brown Hair} &
  \multirow{2}{*}{Glasses} \\ \cline{2-3}
          & Path 1  & Path 2  &         &         &         &         &         &         \\ \hline
Idloss \cite{Deng2019}   & 54.90\% & 57.81\% & 95.63\% & 0.7418 & 0.8466 & 0.6859 & 0.5210 & 0.7231 \\ \hline
No-Idloss & 59.41\% & 60.49\% & 94.97\% & 0.9066 & 0.8935 & 0.8064 & 0.6100 & 0.9265 \\ \hline
\end{tabular}%
}
\label{tab:ablation}
\end{table*}

\begin{figure}[!ht]
\centering
\includegraphics[width=0.45\textwidth]{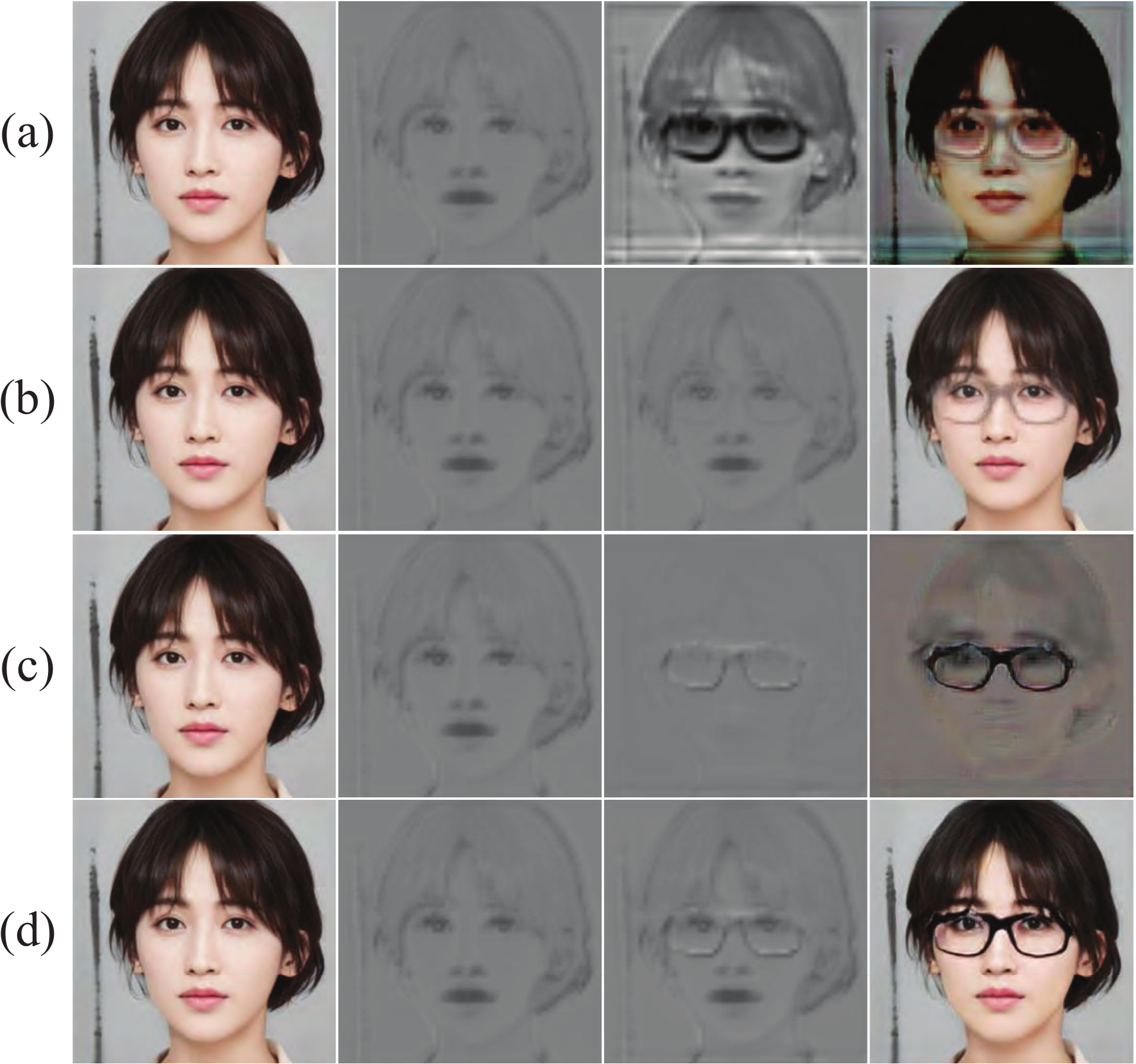}
\caption{The ablation studies for attention and affine transformation modules. Left to right for each line denotes: source image, feature map from encoder, fused feature map, and edited facial image.}
\label{fig:attention}
\end{figure}

We config four feature fusion strategies to investigate attention and affine transformation modules. Fig. \ref{fig:attention}(a) is the experimental result after removing the attention and affine transformation module. From the fused feature map, we can see that some irrelevant regions are activated. It results in a large gap between the source image and generated one. In addition, the desired attribute is not edited to the facial image correctly. Fig. \ref{fig:attention}(b) presents the results that only retain the affine transformation module. It can be seen that the regions unrelated to glasses in the fused feature map are not changed. The results indicate that CCR learns the disentangled style codes, which only contain attributes related to glasses and without affecting other attributes. Fig. \ref{fig:attention}(c) shows the results that only retain the attention module. The fused feature map shows that the regions related to glasses are activated while other regions remain unchanged. The glasses are edited to the generated facial image while other regions are obscured. Fig. \ref{fig:attention}(d) is the results that retain both modules. The fused feature map illustrates that the regions related to glasses are activated while other regions remain unchanged. We can get an edited facial image that wears glasses while hair color, bangs, and facial identity remain unchanged. We experimentally conclude that disentangled style codes and accurate attention regions are necessary condition for editing continuity, consistency, and reversibility.

\section{Conclusion and Discussion}
In this work, we propose a novel method named CCR for facial image editing, which can achieve the goal of editing continuity, consistency, and reversibility. Extensive qualitative and quantitative experimental results demonstrate the effectiveness of the proposed method. Furthermore, we introduce a sufficient criterion, which can be used to determine whether a model is satisfied with continuity, consistency, and reversibility. We experimentally conclude that a generalized encoder, disentangled style codes, and accurate attention regions are the necessary conditions for editing continuity, consistency, and reversibility.

We take a first step towards a continuous, consistent, and reversible model. We believe that our proposed concepts and model will have promising prospects for multimedia applications, and we expect to create a graphical interface program, which can be integrated as a mobile application used in social software.

\bibliographystyle{IEEEtran}
\bibliography{Collection}

\end{document}